\documentclass{article}

\usepackage[final, nonatbib]{neurips_2024}

\usepackage[utf8]{inputenc} % allow utf-8 input
\usepackage[T1]{fontenc}    % use 8-bit T1 fonts
\usepackage[colorlinks=true, linkcolor=red, citecolor=cyan, urlcolor=cyan]{hyperref}
\usepackage{url}            % simple URL typesetting
\usepackage{booktabs}       % professional-quality tables
\usepackage{amsfonts}       % blackboard math symbols
\usepackage{nicefrac}       % compact symbols for 1/2, etc.
\usepackage{microtype}      % microtypography
\usepackage{multirow}
\usepackage{subcaption}
\usepackage[pdftex]{graphicx}
\usepackage{amsmath}
\usepackage{amssymb}
\usepackage{bbm}
\usepackage{breqn}
\usepackage{makecell}
\usepackage{pifont}% http://ctan.org/pkg/pifont
\usepackage{soul}
\usepackage[dvipsnames]{xcolor}

\usepackage{color, colortbl}
\newcommand{\cmark}{\ding{51}}
\newcommand{\xmark}{\ding{55}}
\definecolor{Gray}{gray}{0.9}
\definecolor{NormalGray}{gray}{0.6}
\definecolor{LightCyan}{rgb}{0.88,1,1}

\DeclareMathOperator*{\argmax}{arg\,max}

\title{Unified Speech Recognition: A Single Model for Auditory, Visual, and Audiovisual Inputs}

\author{%
  Alexandros Haliassos \\
  Imperial College\\
  \texttt{ah2214@ic.ac.uk} \\
  \And
  Rodrigo Mira \\
  Imperial College\\
  \texttt{rs2517@ic.ac.uk} \\
  \And
  Honglie Chen \\
  Meta AI \\
  \texttt{hongliechen@meta.com} \\
  \And
  Zoe Landgraf \\
  Meta AI \\
  \texttt{zoelandgraf@meta.com} \\
  \And
  Stavros Petridis \\
  Meta AI / Imperial College \\
  \texttt{stavrosp@meta.com} \\
  \And
  Maja Pantic \\
  Meta AI / Imperial College \\
  \texttt{majapantic@meta.com} \\
}

\begin{document}

\maketitle

\begin{abstract}
Research in auditory, visual, and audiovisual speech recognition (ASR, VSR, and AVSR, respectively) has traditionally been conducted independently. Even recent self-supervised studies addressing two or all three tasks simultaneously tend to yield separate models, leading to disjoint inference pipelines with increased memory requirements and redundancies. This paper proposes unified training strategies for these systems. We demonstrate that training a single model for all three tasks enhances VSR and AVSR performance, overcoming typical optimisation challenges when training from scratch. Moreover, we introduce a greedy pseudo-labelling approach to more effectively leverage unlabelled samples, addressing shortcomings in related self-supervised methods. Finally, we develop a self-supervised pre-training method within our framework, proving its effectiveness alongside our semi-supervised approach. Despite using a single model for all tasks, our unified approach achieves state-of-the-art performance compared to recent methods on LRS3 and LRS2 for ASR, VSR, and AVSR, as well as on the newly released WildVSR dataset. Code and models are available at \url{https://github.com/ahaliassos/usr}.
\end{abstract}

\section{Introduction} \label{sec:intro}
Speech recognition can be achieved using auditory signals (known as auditory/automatic speech recognition; ASR)~\cite{abdel2014convolutional,chiu2018state}, visual cues from lip movements (visual speech recognition; VSR)~\cite{assael2016lipnet,martinez2020lipreading}, or both (audiovisual speech recognition; AVSR)~\cite{petridis2018end,ma2021end}. Audio typically offers the most relevant information in videos of talking faces, but lipreading can greatly enhance recognition, especially when the audio is noisy or wholly unavailable~\cite{ma2021end}. Despite the similarities between ASR, VSR, and AVSR, research in these fields has largely developed independently~\cite{baevski2020wav2vec,hsu2021hubert,assael2016lipnet,ma2022visual}.

The Transformer architecture's versatility~\cite{vaswani2017attention,akbari2021vatt,girdhar2022omnivore} has spurred efforts to unify speech recognition by pre-training a single model on various unlabelled inputs (visual, auditory, and audiovisual) through self-supervision~\cite{shi2022learning,zhu2023vatlm,lian2023av}. 
However, these methods often require separate fine-tuning stages for ASR, VSR, and AVSR, leading to separate models for each task, which increases computational load and complexity. u-HuBERT~\cite{hsu2022u} shows that a single pre-trained model \textit{can} be fine-tuned for all three tasks, yet does not reach the performance of separately fine-tuned models~\cite{haliassos2022jointly,haliassos2024braven}.

In this paper, we delve deeper into strategies for unified speech recognition (USR) by training a single model to perform ASR, VSR, and AVSR. We find that training such a model \textit{from scratch} on the LRS3 dataset~\cite{afouras2018lrs3} achieves competitive performance on all tasks. This is notable given the known optimisation difficulties in VSR training, which previously required self-supervised pre-training~\cite{shi2022learning}, supervised feature extractor pre-training~\cite{ma2021end}, or curriculum learning strategies~\cite{ma2022visual}. Our findings suggest that including audio improves the optimisation landscape for VSR and AVSR supervised training, as observed in a different context by~\cite{djilali2023lip2vec}.

Furthermore, we propose a semi-supervised pseudo-labelling approach to leverage unlabelled audiovisual data, addressing shortcomings of standard fine-tuning in self-supervised methods~\cite{shi2022learning,zhu2023vatlm,haliassos2022jointly,haliassos2024braven}. Fine-tuning often leads to overfitting due to using fewer samples than pre-training, requiring various ``tricks'' to reach optimal performance~\cite{shi2022learning,haliassos2022jointly}. This issue is particularly pronounced in encoder-decoder architectures where usually only the encoder is pre-trained, and attempts to pre-train the decoder have yielded inconsistent results~\cite{ao2022pre,elkahky2023coarser}. Our semi-supervised approach generates pseudo-labels via an encoder-decoder momentum-based teacher~\cite{tarvainen2017mean} to leverage unlabelled samples throughout training, effectively mitigating overfitting. Training on all three modalities simultaneously helps alleviate the computational cost of pseudo-labelling as the cost is amortised across the inputs.

Lastly, inspired by recent self-supervised works, we design a pre-training method within our unified framework. We combine pre-training with pseudo-labelling and show that our semi-supervised approach is complementary to self-supervision. Our final unified models achieve state-of-the-art results across multiple settings, surpassing existing methods that use separate models for each task. 

\vspace{-0.1cm}
\section{Related Work} \label{sec:related_work}
\paragraph{Audiovisual self-supervised speech representation learning.} 

Recent interest in audiovisual self-supervised learning for speech recognition has focused on leveraging the correspondence between audio waveforms and silent lip movements to capture shared semantic content across the modalities~\cite{shi2022learning,haliassos2022jointly,zhu2023vatlm,lian2023av,haliassos2024braven}. These methods employ cross-modal learning and masked prediction~\cite{devlin2018bert} to develop contextualised representations from large unlabelled datasets, which are more readily available than transcribed datasets. After pre-training, a randomly initialised decoder is appended to the encoder, often with an optional CTC layer~\cite{graves2006connectionist}. The system is then fine-tuned on a smaller set of labelled samples for tasks such as ASR, VSR, and AVSR, usually resulting in different models for each task~\cite{shi2022learning,lian2023av}. However, these methods may fail to leverage unlabelled samples fully since the pretext tasks are not directly aligned with speech recognition. Furthermore, the decoder, trained on limited data during fine-tuning, is highly susceptible to overfitting, necessitating strategies such as freezing encoder layers~\cite{shi2022learning} or employing variable learning rates across layers to optimise performance~\cite{haliassos2022jointly,clark2020electra}.

\paragraph{Pseudo-labelling for speech recognition.}
Pseudo-labelling has been explored in audiovisual speech recognition literature, with methods such as offline pseudo-labelling~\cite{ma2022visual,ma2023auto} and frame-wise distillation using frozen teacher models~\cite{afouras2020asr}. While these approaches rely on frozen external ASR models trained on large-scale datasets~\cite{baevski2020wav2vec,panayotov2015librispeech}, our USR method eliminates this dependency using a randomly initialised teacher model that improves throughout training.

\textit{Iterative} pseudo-labelling has shown promise for ASR. Some employ multiple rounds of pseudo-labelling using costly beam search and filtering strategies~\cite{park2020improved,xu2020iterative,zhang2020pushing,kahn2020self}, while others continuously and efficiently update pseudo-labels using a CTC-only loss~\cite{likhomanenko2020slimipl,higuchi2021momentum}. However, eliminating filtering and attention losses can impact training due to low-quality pseudo-labels, as observed in a recent method~\cite{rouditchenko2023av} that aims to apply these approaches for ASR, VSR, and AVSR but lags behind the state-of-the-art (see Appendix \ref{app:avcpl_comparison}). In contrast, USR uses an encoder-decoder architecture to generate CTC and attention pseudo-labels at each iteration through a greedy approach, while pseudo-label quality is maintained via a token-wise filtering mechanism inspired by the semi-supervised FixMatch technique~\cite{sohn2020fixmatch} in image recognition. We note that sharing the same pseudo-labels across auditory, visual, and audiovisual inputs amortises generation costs, leading to efficient CTC-attention training. 

\paragraph{Single model for multiple modalities.} An earlier study~\cite{makino2019recurrent} trained a single recurrent neural network~\cite{sherstinsky2020fundamentals} for ASR, VSR, and AVSR, but noted significant performance differences compared to modality-specific models. Recent works have shown that the Transformer architecture~\cite{vaswani2017attention} can handle multiple modalities using the same weights, with minimal performance degradation~\cite{akbari2021vatt,girdhar2022omnivore}. In speech recognition, some~\cite{shi2022learning,zhu2023vatlm,lian2023av} use the same Transformer encoder for auditory, visual, and audiovisual inputs during pre-training, but then separately fine-tune the parameters for ASR, VSR, and AVSR, resulting in separate models during deployment. u-HuBERT~\cite{hsu2022u} uses the same weights for all modalities when fine-tuning a pre-trained AV-HuBERT backbone~\cite{shi2022learning}, demonstrating the viability of a unified model. However, it encounters limitations common to other self-supervised approaches, such as proneness to overfitting during supervised fine-tuning. Our proposed semi-supervised approach leverages unlabelled samples during the fine-tuning stage, significantly alleviating these concerns.

\section{Unified Speech Recognition} \label{sec:method}
\begin{figure}
  \centering
  \includegraphics[width=\linewidth]{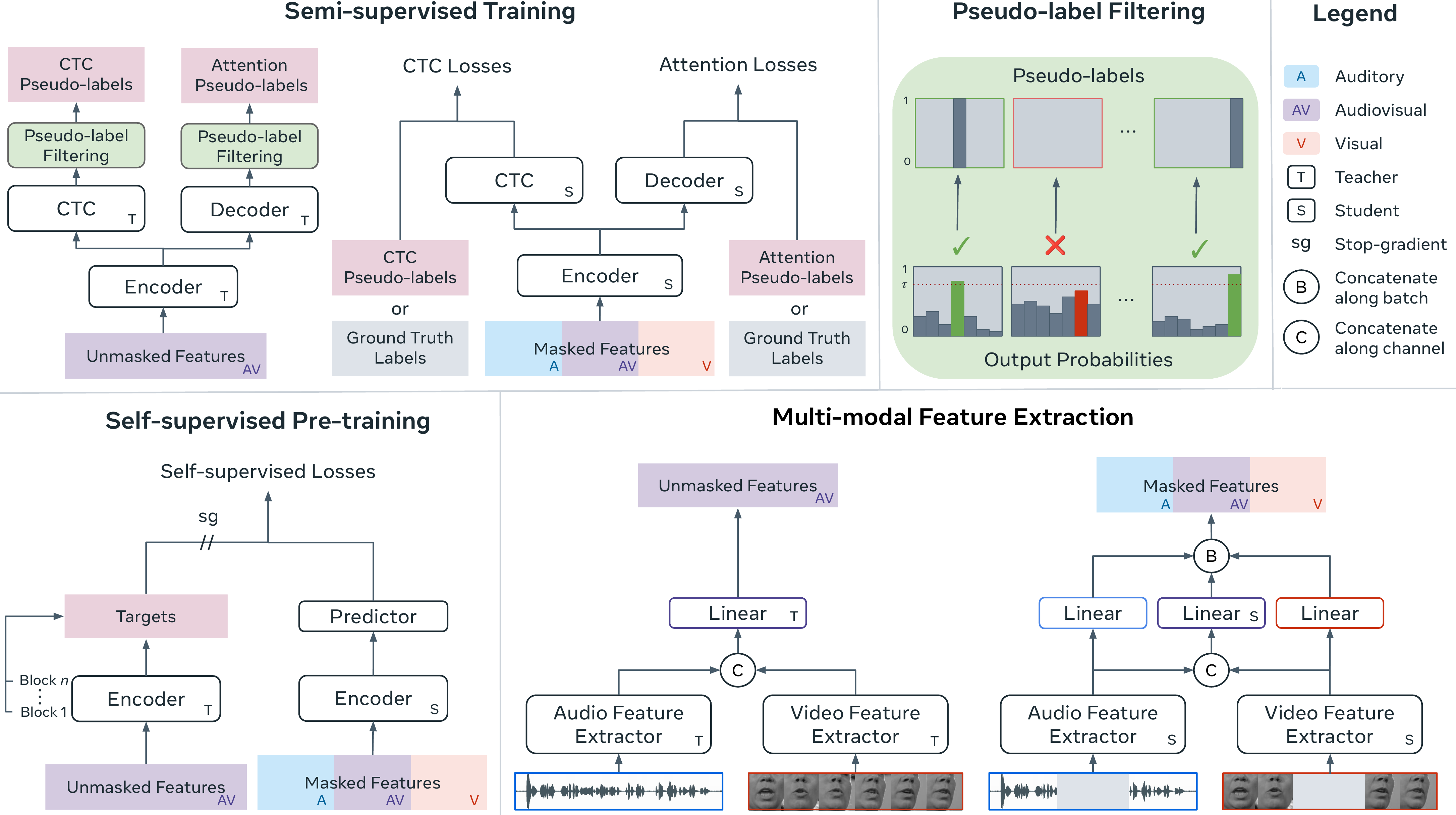}
  \caption{\textbf{Unified Speech Recognition.} Our USR method combines self-supervised pre-training with semi-supervised fine-tuning. For \textbf{semi-supervised training}, pseudo-labels are generated from unmasked audiovisual features using an EMA (exponential moving average)-based teacher. The student, intaking masked inputs, predicts pseudo-labels for unlabelled data and ground-truth labels for labelled data. \textbf{To obtain the pseudo-labels}, an argmax operation is applied to the CTC and attention teacher output probabilities; the tokens with predicted probability below a fixed threshold are discarded. For \textbf{self-supervised pre-training}, a student encoder processes masked visual, auditory, and audiovisual samples and predicts targets, generated by an EMA-based teacher intaking unmasked audiovisual samples, via a shallow predictor. The targets are the average outputs of the teacher blocks. The resulting student weights are used to initialise the student and teacher in semi-supervised fine-tuning. \textbf{Feature extraction} is achieved through modality-specific feature extractors, whose features are concatenated along the channel dimension to produce the audiovisual inputs. The auditory, visual, and audiovisual student inputs are batched together for training efficiency.}
  \label{fig:overview}
\end{figure}

Our unified method trains a pre-LN~\cite{xiong2020layer} Transformer~\cite{vaswani2017attention} encoder-decoder model for ASR, VSR, and AVSR. Section~\ref{sec:scratch} describes the task of unified speech recognition using supervised training, where we have ground-truth annotation for each audio-visual pair. Sections~\ref{sec:semi} and \ref{sec:self} then introduce our proposed idea, which employs semi-supervised training and self-supervised pre-training to effectively utilise unlabelled samples. An overview of USR's components is depicted in Figure~\ref{fig:overview}.

\subsection{Unified Supervised Training} \label{sec:scratch}

\paragraph{Inputs.} Let $\{(\mathbf{v}_b,\mathbf{a}_b,\mathbf{y}_b):b\in [1,B]\}$ be a batch of $B$ labelled samples, where $\mathbf{v}_b$ denotes a $T_{\text{v}}$-frame video of lip movements, $\mathbf{a}_b$ denotes the corresponding (raw) audio waveform of $T_{\text{a}}=640T_{\text{v}}$ frames\footnote{We assume the video is sampled at 25 frames per second and the audio at 16,000kHz.}, and $\mathbf{y}_b$ denotes the label sequence of length $T_{\text{l}}$. Following~\cite{ma2022visual,haliassos2022jointly}, $\mathbf{v}_b$ and $\mathbf{a}_b$ are zero-masked with a maximum duration of 0.4 and 0.6 seconds for each second of video and audio, respectively.

\paragraph{Multi-modal feature extraction.} The raw video and audio are fed into ResNet-18~\cite{he2016deep} architectures: a 2D version with a 3D stem ~\cite{stafylakis2017combining} for video and a 1D version for audio, sub-sampling the audio to match the video's sampling rate~\cite{haliassos2022jointly}. Linear layers follow the feature extractors to produce the visual and auditory features. The audiovisual features are formed by concatenating the feature extractor outputs along the channel dimension and applying a linear transformation. Finally, the features from the three modalities are concatenated along the batch dimension for efficient processing. We provide the model with all three input types, enabling it to perform well on ASR, VSR, and AVSR.

\paragraph{Losses.} The encoder outputs pass through a linear + softmax layer to yield output probabilities $\mathbf{c}_{b,m}$ for each modality $m\in \{\text{v},\text{a},\text{av}\}$. The CTC loss for each modality is given by 
\begin{align}
    \mathcal{C}_m=\frac{1}{B}\sum_{b=1}^{B}l_{\text{ctc}}(\mathbf{c}_{b,m},\mathbf{y}_b),
\end{align}
where $l_{\text{ctc}}$ is the standard CTC loss~\cite{graves2006connectionist}. Further, let $\mathbf{a}_{b,m}$ denote the attention probabilities from the outputs of the decoder in \textit{teacher forcing} mode~\cite{williams1989learning}. The batch attention loss can be expressed as 
\begin{align}
    \mathcal{A}_m=\frac{1}{B}\sum_{b=1}^Bl_{\text{ce}}(\textbf{a}_{b,m},\textbf{y}_b), 
\end{align}
where $l_{\text{ce}}$ is the summed cross-entropy loss for each token. The CTC and attention losses are combined to obtain
\begin{align}  \mathcal{L}_m=\lambda_{\text{ctc}}\mathcal{C}_m+(1-\lambda_{\text{ctc}})\mathcal{A}_m, \label{eq:lab}
\end{align}
where $\lambda_{\text{ctc}}$ is the relative weight placed on the CTC loss versus the attention loss. We set $\lambda_{\text{ctc}}$ to 0.1, following \cite{ma2023auto,haliassos2022jointly,haliassos2024braven}.  The overall labelled loss is given by
\begin{align} \label{eq:scratch_final_loss}
    \mathcal{L}^{\text{lab}}=\lambda_{\text{v}} \mathcal{L}_{\text{v}} + (1-\lambda_{\text{v}}) (\mathcal{L}_{\text{a}} + \mathcal{L}_{\text{av}}),
\end{align}
where $\lambda_{\text{v}}$ controls the weight of the video loss relative to the audio/audiovisual losses. We do not use separate weights for the audio/audiovisual losses due to similar training dynamics observed in preliminary experiments.

\subsection{Unified Semi-supervised Training} \label{sec:semi}

We introduce a student-teacher pseudo-labelling framework to utilise \textit{unlabelled} samples alongside labelled examples. The student, equipped with labelled losses, mirrors the model in Section~\ref{sec:scratch}.

\paragraph{Inputs.} In addition to the labelled batch from Section~\ref{sec:scratch}, we now also have $B^{\text{u}}$ \textit{unlabelled} video and audio samples $\{(\mathbf{v}_b^{\text{u}},\mathbf{a}_b^{\text{u}}):b\in [1,B^{\text{u}}]\}$. The student inputs are masked as before.

\paragraph{Pseudo-labels.} The teacher, sharing the same architecture as the student, generates pseudo-labels for unlabelled samples. The student is optimised as usual, but no gradients are passed to the teacher. Instead, the teacher's weights $\theta_{t}$ are updated at each iteration via an exponential moving average (EMA) of the student's weights $\theta_{s}$~\cite{grill2020bootstrap}: $\theta_{t}\leftarrow\mu \theta_{t}+(1-\mu)\theta_{s}$, where $\mu$ increases throughout training from 0.999 to 1 using a cosine scheduler.

For an unmasked audiovisual sample, let $\mathbf{\tilde{c}}_b$ and $\mathbf{\tilde{a}}_b$ denote the CTC probabilities from the teacher encoder and the attention probabilities from the teacher decoder, respectively. The CTC and attention pseudo-labels are given by $\argmax(\mathbf{\tilde{c}}_b)$ and $\argmax(\mathbf{\tilde{a}}_b)$, respectively, where $\argmax$ is applied token-wise. Hence, the pseudo-labels correspond to units with the maximum probability across the vocabulary for each input/output time-step. The attention targets are generated auto-regressively by selecting, at each time-step, the most likely unit as the input for the next time-step, without using a costly beam search strategy. Our greedy approach allows for efficient label generation.

\paragraph{Filtering.} The teacher may not consistently generate high-quality predictions, especially early in training. We propose a straightforward token-wise filtering mechanism, creating masks $\mathbbm{1}(\max(\mathbf{\tilde{c}}_b)\geq\tau)$ and $\mathbbm{1}(\max(\mathbf{\tilde{a}}_b)\geq\tau)$, where the operations are applied token-wise. We thus discard a pseudo-label for a given time-step if its confidence falls below a certain threshold $\tau$. This mechanism draws inspiration from image recognition literature~\cite{sohn2020fixmatch} and is adapted to sequences.

\paragraph{Unlabelled losses.} The unlabelled losses are computed via the cross-entropy between the student predictions and the teacher pseudo-labels. That is, the per-modality CTC losses are given by 
\begin{align}
    \mathcal{C}_m^{\text{u}}=\frac{1}{B^{\text{u}}}\sum^{B^{\text{u}}}_{b=1}\mathbbm{1}(\max(\mathbf{\tilde{c}}_b)\geq\tau)\odot l_{\text{ce}}(\textbf{c}_{b,m}^{\text{u}},\argmax(\mathbf{\tilde{c}}_b)),
\end{align}
where $\odot$ denotes the Hadamard product and $\textbf{c}_{b,m}^{\text{u}}$ the student outputs. The attention losses $\mathcal{A}_{m}^{\text{u}}$ are computed similarly. The unlabelled losses $\mathcal{L}_m^{\text{u}}$ are obtained as in Eq. \ref{eq:lab}:
\begin{align}
    \mathcal{L}_m^{\text{u}}=\lambda_{\text{ctc}}\mathcal{C}_m^{\text{u}}+(1-\lambda_{\text{ctc}})\mathcal{A}_{m}^{\text{u}},
    \label{eq:unlab}
\end{align}

\paragraph{Final loss.} The total semi-supervised loss $\mathcal{L}^{\text{semi}}$ combines the per-modality labelled (see Eq.~\ref{eq:lab}) and unlabelled losses (see Eq.~\ref{eq:unlab}):
\begin{dmath}    \mathcal{L}^{\text{semi}}=\gamma_{\text{v}}\lambda_{\text{v}}\mathcal{L}_{\text{v}}+\gamma_{\text{a}} (1-\lambda_{\text{v}}) (\mathcal{L}_{\text{a}}+\mathcal{L}_{\text{av}})+(1-\gamma_{\text{v}})\lambda_{\text{v}}\mathcal{L}_{\text{v}}^{\text{u}}+(1-\gamma_{\text{a}}) (1-\lambda_{\text{v}}) (\mathcal{L}_{\text{a}}^{\text{u}}+\mathcal{L}_{\text{av}}^{\text{u}}), \label{eq:final_loss}
\end{dmath}
where $\gamma_{\text{a}}$ and $\gamma_{\text{v}}$ weigh the contribution of the labelled loss versus the unlabelled loss for audio/audiovisual and visual inputs, respectively. In Section \ref{sec:semi_properties}, we show the benefits of using separate weights for each modality rather than a single weight for both.

\subsection{Unified Self-supervised Pre-training} \label{sec:self}

Transformers typically benefit from self-supervised pre-training~\cite{he2022masked, shi2022learning, haliassos2022jointly, lian2023av}, even with the same data used during fine-tuning~\cite{caron2021emerging,he2022masked}. Inspired by recent work~\cite{haliassos2022jointly,haliassos2024braven,lian2023av}, we propose a self-supervised method within our framework that can precede semi-supervised fine-tuning.

\paragraph{Inputs.} For pre-training, we use only the unlabelled $B^u$ samples from Section~\ref{sec:semi}. Following~\cite{haliassos2022jointly}, we mask the student inputs by selecting each video frame index as the start of a three-frame mask with a 0.4 probability, applying a corresponding enlarged mask to the audio in temporal alignment. The elements of the mask $\mathbf{h}_b$ are set to 0 and 1 for unmasked and masked tokens, respectively.

\paragraph{Targets.} The targets are generated by an EMA-based teacher encoder model from unmasked \textit{audiovisual} inputs, similarly to Section~\ref{sec:semi}. Following~\cite{lian2023av,haliassos2024braven}, the targets $\textbf{e}_b$ are generated by averaging the outputs from all encoder blocks and applying instance normalisation ~\cite{ulyanov2016instance}. Using only audio targets, as in~\cite{lian2023av}, can make the student's final layers more relevant to speech, which has proven beneficial for fine-tuning with few samples, where there is high chance of overfitting~\cite{haliassos2022jointly}. Our fine-tuning process instead uses abundant unlabelled data with pseudo-labels which help reduce overfitting and allow the network to learn from rich audiovisual targets.

\paragraph{Predictor.} Following~\cite{haliassos2022jointly}, we employ a 512-dimensional two-block Transformer predictor that processes student encoder outputs and mask tokens to produce predictions $\textbf{p}_{b,m}$. Unlike the separate predictors for video and audio used in~\cite{haliassos2022jointly}, we use a single predictor for all inputs.

\paragraph{Loss.} The loss for modality $m$ can be expressed as
\begin{align}
    \mathcal{L}_m^{\text{self}}=-\frac{1}{B^{\text{u}}}\sum^{B^{\text{u}}}_{b=1}\mathbf{h}_b\odot \cos(\mathbf{p}_{b,m}, \mathbf{e}_b),
\end{align}
where $\cos$ denotes cosine similarity, applied token-wise. Thus, the student aims to predict the teacher targets corresponding to the masked inputs. The self-supervised loss $\mathcal{L}^{\text{self}}$ is then
\begin{align}
    \mathcal{L}^{\text{self}}=\lambda_{\text{v}} \mathcal{L}_{\text{v}}^{\text{self}} + (1-\lambda_{\text{v}}) (\mathcal{L}_{\text{a}}^{\text{self}} + \mathcal{L}_{\text{av}}^{\text{self}}).
\end{align}

\vspace{-0.1cm}
\section{Main Properties} \label{sec:main_properties}
In this section, we investigate the behaviour of our unified model. For all experiments, we use a 12-block Base model with hidden size of 512 (see Appendix~\ref{app:model_config} for model details). We report test set word error rates (WER) for direct comparison with the main results. Note that we used the validation set from~\cite{shi2022learning} in the exploration stage to avoid overfitting to the test set.

\subsection{Unified Supervised Training} \label{sec:scratch_properties}

\begin{table}
\caption{\textbf{Supervised ablations} on the full LRS3 dataset using our Base model. Default settings are in \colorbox{Gray}{gray} in all tables of the paper.}
\vspace{-0.2cm}
\label{table:scratch}
\begin{subtable}[h]{0.33\textwidth}
\centering
\caption{\textbf{Sharing model parameters vs. using modality-specific models.}}
\begin{tabular}[b]{l c c c}\toprule
\multirow{2}{*}{Params} & \multicolumn{3}{c}{WER (\%)} \\  \cmidrule(lr){2-4}
& V & A & AV \\ \midrule
\rowcolor{Gray}
Shared & \textbf{36.4} & 2.3 & \textbf{2.1} \\
Unshared & 85.5 & \textbf{2.1} & 63.4 \\
\bottomrule 
\end{tabular}
\label{table:sharing_params}
\end{subtable}
\hfill
\begin{subtable}[h]{0.28\textwidth}
\centering
\caption{\textbf{Modality sampling.} Random sampling is trained for $3\times$ more epochs as it sees one-third of the data at each iteration.}
\begin{tabular}[b]{l c c c}\toprule
\multirow{2}{*}{Mod} & \multicolumn{3}{c}{WER (\%)} \\  \cmidrule(lr){2-4}
 & V & A & AV \\ \midrule
Rand & \textbf{36.2} & \textbf{2.3} & 2.2 \\
\rowcolor{Gray}
All & 36.4 & \textbf{2.3} & \textbf{2.1} \\
\bottomrule 
\end{tabular}
\label{table:modality_sampling}
\end{subtable}
\hfill
\begin{subtable}[h]{0.28\textwidth}
\vspace{-0.35cm}
\centering
\caption{\textbf{Relative weight for video loss.}}
\begin{tabular}[b]{l c c c}\toprule
\multirow{2}{*}{$\lambda_v$} & \multicolumn{3}{c}{WER (\%)} \\  \cmidrule(lr){2-4}
 & V & A & AV \\ \midrule
0.1 & 42.9 & \textbf{2.2} & \textbf{1.9} \\
\rowcolor{Gray}
0.3 & 36.4 & 2.3 & 2.1 \\
0.5 & \textbf{35.2} & 2.4 & 2.2 \\
\bottomrule 
\end{tabular}
\label{table:relative_video_weight}
\end{subtable}
\end{table}

In Table~\ref{table:scratch}, we investigate properties of training our unified model from scratch on the full LRS3 dataset~\cite{afouras2018lrs3} (see Section~\ref{sec:scratch}). Training details are provided in Section~\ref{app:lab_settings}.

\paragraph{Sharing weights.} Table~\ref{table:sharing_params} studies the impact of weight sharing versus separate models per task (ASR, VSR, AVSR). While using only auditory inputs yields strong performance, training VSR and AVSR models from scratch encounters optimisation challenges, in line with prior research~\cite{shi2022learning,haliassos2022jointly}. Interestingly, these hurdles are overcome with weight sharing, resulting in robust VSR and AVSR performance without self-supervised pre-training~\cite{haliassos2022jointly} or training techniques like curriculum learning~\cite{ma2022visual}. This is likely due to audio containing denser verbal information than video, enhancing the optimisation landscape for visual modalities~\cite{djilali2023lip2vec}.

\paragraph{Modality sampling.} We employ a weighted average to combine the per-modality losses (see Eq.~\ref{eq:scratch_final_loss}). In contrast, other methods~\cite{shi2022learning,lian2023av} randomly sample, at each iteration, input types with different probabilities, which may vary during training. Table~\ref{table:modality_sampling} shows that our approach performs similarly with random sampling when training the latter for $3\times$ more epochs. Our approach offers benefits such as sharing computational costs among feature extractor forward passes and amortising the cost of pseudo-label generation across input types (see Section~\ref{sec:semi}), as all modalities use the same targets.

\paragraph{Input type weight.} Table~\ref{table:relative_video_weight} studies the effect of using different weights for the visual modality. We observe that using a higher $\lambda_{\text{v}}$ for the VSR loss improves VSR but worsens ASR/AVSR. We choose $\lambda_{\text{v}}=0.3$ as our default setting, striking a balance in performance among the different tasks. 

\subsection{Unified Semi-supervised Training} \label{sec:semi_properties}

In Table~\ref{fig:unlabelled_data}, we ablate various components to better understand our unified semi-supervised framework (see Section~\ref{sec:self}). We adopt the common low-resource setting~\cite{shi2022learning}: the 30-hour ``trainval'' partition of LRS3 serves as our labelled dataset, while the remaining portion of LRS3 (without labels) provides our unlabelled samples. See Appendix~\ref{app:lab_settings} for training details.

\begin{figure}[t]
    \centering
    % First row with a subfigure and a subtable
    \begin{subfigure}[t]{0.7\textwidth}
        \centering
        \includegraphics[width=\linewidth]{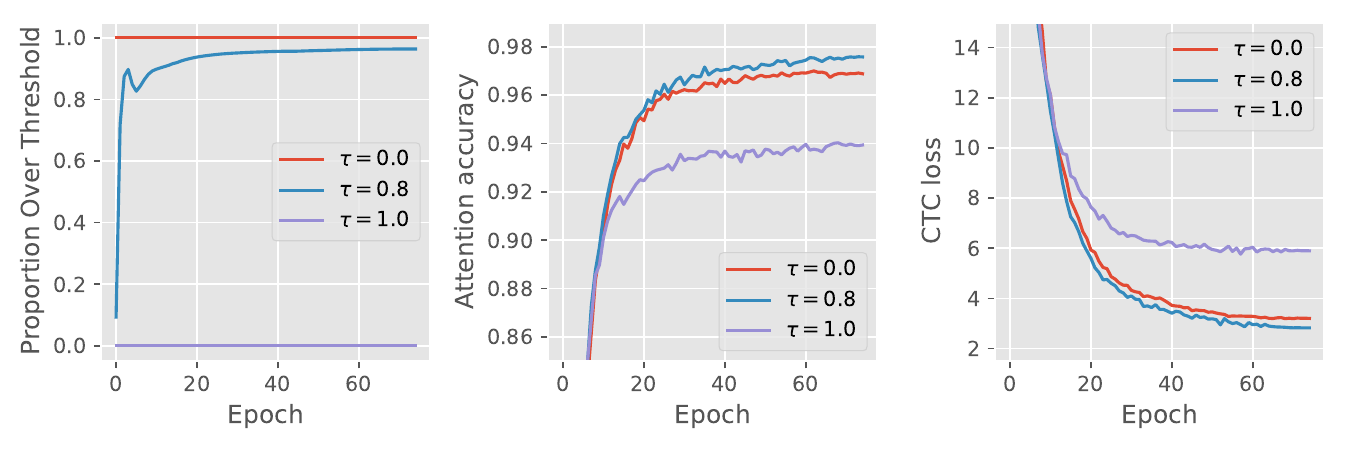}
        \vspace{-0.5cm}
    \end{subfigure}%
    \raisebox{1.5cm}{
    \begin{subtable}[t]{0.3\textwidth}
        \centering
        \resizebox{0.9\linewidth}{!}{
        \begin{tabular}{l c c c}\toprule
        \multirow{2}{*}{$\tau$} & \multicolumn{3}{c}{WER (\%)} \\
        \cmidrule(lr){2-4}
        & V & A & AV \\
        \midrule
         0.0 & 40.7 & 4.9 & 4.7 \\
         \rowcolor{Gray}
         0.8 & \textbf{37.8} & \textbf{4.0} & \textbf{3.9} \\ 
         1.0 & 61.8 & 8.9 & 8.4 \\
        \bottomrule 
        \end{tabular}
        }
    \end{subtable}}
    \vspace{-0.5cm}
    \caption{\textbf{Pseudo-label filtering threshold.} \textbf{Left}: Validation plots for different values of threshold $\tau$. \textbf{Right}: Final WER for different values of $\tau$.}
    \label{fig:filtering_threshold}
\end{figure}

\paragraph{Filtering predicted tokens.} Figure~\ref{fig:filtering_threshold} investigates the impact of the threshold parameter $\tau \in \{0, 0.8, 1\}$. We plot (from left to right) (1) the proportion of tokens exceeding $\tau$, (2) the validation attention accuracy of the decoder using teacher forcing, and (3) the CTC loss, as a function of the epoch number. We also show the final WER. We observe that $\tau=1$, where only labelled samples contribute to training, results in poor attention accuracy, high CTC loss, and high WER across input types. Conversely, $\tau=0$, implying no filtering (\textit{i.e.}, all tokens are considered regardless of confidence level), yields competitive performance, suggesting some robustness to low-quality pseudo-labels. Finally, for $\tau=0.8$, the proportion of tokens with confidence over $\tau$ begins at a low level and steadily increases throughout training as the teacher network improves. This yields improved performance in terms of attention accuracy, CTC loss, and final WER, demonstrating the efficacy of filtering via a simple confidence threshold. A more fine-grained analysis of $\tau$ values are given in Section \ref{sec:more_semi_ablations}.

\paragraph{Quantity/quality trade-off.} Pseudo-labels tend to be abundant but noisy, while ground-truth transcriptions are scarce yet high-quality. The balance between quantity and quality is adjustable via the hyperparameters $\gamma_{\text{v}}$ and $\gamma_{\text{a}}$ in Eq.~\ref{eq:final_loss}. Table~\ref{table:relative_weights} explores different values for $\gamma_{\text{v}}$ and $\gamma_{\text{a}}$, revealing better performance when $\gamma_{\text{a}}>\gamma_{\text{v}}$. Noisy pseudo-labels generated from audiovisual samples may suffice for VSR, which often performs worse than ASR/AVSR and benefits from data abundance. Conversely, ASR/AVSR is less prone to overfitting and may suffer with excessive reliance on low-quality pseudo-labels, requiring a higher relative weight on labelled losses.

\paragraph{Momentum.} Table~\ref{table:momentum} shows the effect of updating the teacher's weight via EMA ($\mu=0.999$) compared to simply copying the student's weights at every iteration ($\mu=0$). Using EMA results in better performance, yet good results are achieved even without it.

\paragraph{Loss types.} CTC and attention-based encoder-decoder frameworks are dominant approaches in speech recognition. While attention typically outperforms CTC, it may struggle with proper alignment prediction, requiring tuning of various decoding hyperparameters~\cite{kim2017joint}. To address these challenges, we adopt a CTC-attention hybrid framework~\cite{kim2017joint}, as in~\cite{haliassos2022jointly,ma2022visual,ma2023auto}. The costly auto-regressive attention pseudo-label generation is made computationally feasible via our greedy strategy and multi-modal feature extraction (which amortises pseudo-label generation costs). Table~\ref{table:loss_types} demonstrates a significant improvement in results by using both CTC and attention compared to CTC alone.

\begin{table}[t]
\centering
    \caption{\textbf{Semi-supervised ablations} under the LRS3 low-resource setting using our Base model.}
    \label{fig:unlabelled_data}
    \begin{subtable}[t]{0.32\textwidth}
        \centering
        \caption{\textbf{Relative labelled weight for \linebreak audio and video.}}
        \label{table:relative_weights}
        \begin{tabular}{l l c c c}\toprule
        \multirow{2}{*}{$\gamma_{\text{a}}$} & \multirow{2}{*}{$\gamma_{\text{v}}$} & \multicolumn{3}{c}{WER (\%)} \\
        \cmidrule(lr){3-5}
        & & V & A & AV \\
        \midrule
        0.5 & 0.5 & 42.3 & 4.1 & 4.0 \\
        0.2 & 0.2 & 38.0 & 4.2 & 4.1 \\
        \rowcolor{Gray}
        0.5 & 0.2 & \textbf{37.8} & \textbf{4.0} & \textbf{3.9} \\
        \bottomrule 
        \end{tabular}
    \end{subtable}%
    \hfill
    \begin{subtable}[t]{0.32\textwidth}
        \centering
        \caption{\textbf{Teacher's EMA momentum parameter.}}
        \label{table:momentum}
        \begin{tabular}{l c c c}\toprule
        \multirow{2}{*}{$\mu$} & \multicolumn{3}{c}{WER (\%)} \\
        \cmidrule(lr){2-4}
         & V & A & AV \\
        \midrule
        0 & 38.9 & 4.1 & 4.0 \\
        \rowcolor{Gray}
        0.999 & \textbf{37.8} & \textbf{4.0} & \textbf{3.9} \\
        \bottomrule 
        \end{tabular}
    \end{subtable}%
    \hfill
    \begin{subtable}[t]{0.32\textwidth}
        \centering
        \caption{\textbf{CTC vs. CTC-attention losses.}}
        \label{table:loss_types}
        \begin{tabular}{l c c c}\toprule
        \multirow{2}{*}{Loss type} & \multicolumn{3}{c}{WER (\%)} \\
        \cmidrule(lr){2-4}
         & V & A & AV \\
        \midrule
        CTC & 45.6 & 5.2 & 5.0 \\
        \rowcolor{Gray}
        CTC-att & \textbf{37.8} & \textbf{4.0} & \textbf{3.9} \\
        \bottomrule 
        \end{tabular}
    \end{subtable}
\end{table}

\subsection{Unified Self-supervised Pre-training} \label{sec:self_properties}
Table~\ref{table:self_supervised} examines the main properties of our self-supervised method (see Section~\ref{sec:self}). We fine-tune pre-trained models with different hyperparameters using our semi-supervised approach (Section~\ref{sec:semi}). We use the LRS3 low-resource setting, as in Section~\ref{sec:semi_properties}. See Appendix~\ref{app:self_settings} for training details.

\paragraph{Target modality.} 
In Table~\ref{table:target_type}, we evaluate our method with targets derived from the different input modalities. Across all cases, pre-training outperforms training from scratch, highlighting the complementarity of semi- and self-supervised training. Visual targets enhance VSR but diminish ASR/AVSR performance compared to auditory targets; overall, audiovisual targets consistently perform best. These results suggest that cross-modal-only pre-training may lose crucial modality-specific information, reducing generalisation when fine-tuning \textit{on all data} (including unlabelled samples), \textit{i.e.}, via pseudo-labelling. Our observations are in contrast to previous findings with supervised fine-tuning, where visual or audiovisual pre-training targets tend to underperform~\cite{shi2022learning,haliassos2022jointly,lian2023av}. See Appendix~\ref{app:sup_vs_semi} for an in-depth analysis comparing supervised and semi-supervised fine-tuning. 

\paragraph{Averaging targets.} \cite{lian2023av,haliassos2024braven} demonstrate that using the average of encoder blocks as targets outperforms using the last block alone. Table~\ref{table:averaging_targets} confirms this finding in our setting.

\paragraph{Predictor depth.} In Table~\ref{table:pred_depth}, we study the influence of predictor depth. A deeper predictor yields more abstract encoder representations, while a shallower one retains more task-specific features~\cite{haliassos2022jointly}. We observe strong performance at our default depth of 2. Notably, our semi-supervised fine-tuning approach is less sensitive to predictor depth than standard methods~\cite{haliassos2022jointly,haliassos2024braven}.

\begin{table}[t]
\caption{\textbf{Self-supervised ablations} under the LRS3 low-resource setting using our Base model.}
\begin{subtable}[h]{0.29\textwidth}
\centering
\caption{\textbf{Target type.} ``Scratch'' refers to semi-supervised training only.}
\label{table:target_type}
\begin{tabular}[b]{l c c c}\toprule
\multirow{2}{*}{Target} & \multicolumn{3}{c}{WER (\%)} \\  \cmidrule(lr){2-4}
& V & A & AV \\ \midrule
\textcolor{NormalGray}{Scratch} & \textcolor{NormalGray}{37.8} & \textcolor{NormalGray}{4.0} & \textcolor{NormalGray}{3.9} \\ \midrule
V & 36.2 & 3.7 & 3.4 \\
A & 37.3 & \textbf{3.2} & 3.1 \\
\rowcolor{Gray}
AV & \textbf{36.0} & \textbf{3.2} & \textbf{3.0} \\
\bottomrule 
\end{tabular}
\end{subtable}
\hfill
\begin{subtable}[h]{0.33\textwidth}
\centering
\caption{\textbf{Averaging blocks vs. using only last encoder block.}}
\label{table:averaging_targets}
\begin{tabular}[b]{l c c c}\toprule
\multirow{2}{*}{Target} & \multicolumn{3}{c}{WER (\%)} \\  \cmidrule(lr){2-4}
 & V & A & AV \\ \midrule
Last block & 37.2 & 3.4 & 3.1 \\
\rowcolor{Gray}
Avg blocks & \textbf{36.0} & \textbf{3.2} & \textbf{3.0} \\
\bottomrule 
\end{tabular}
\end{subtable}
\hfill
\begin{subtable}[h]{0.28\textwidth}
\centering
\caption{\textbf{Predictor depth.}}
\label{table:pred_depth}
\begin{tabular}[b]{l c c c}\toprule
\multirow{2}{*}{Depth} & \multicolumn{3}{c}{WER (\%)} \\  \cmidrule(lr){2-4}
 & V & A & AV \\ \midrule
1 & 37.0 & 3.2 & 3.0 \\
\rowcolor{Gray}
2 & \textbf{36.0} & 3.2 & 3.0 \\
4 & 36.9 & \textbf{3.1} & \textbf{2.9} \\
\bottomrule 
\end{tabular}
\end{subtable}
\label{table:self_supervised}
\end{table}

\vspace{-0.1cm}
\section{Comparisons with Previous Results} \label{sec:sota_comparisons}
\begin{table}
\centering
\caption{\textbf{Comparisons with self-supervised methods.} LRS3 results for the low-resource (LR) and high-resource (HR) labelled data settings, with 30 and 433 hours of labelled data, respectively. Best results in \textbf{bold}, second-best \underline{underlined}.}
\label{table:self_supervised_lrs3}
\vspace{0.1cm}
\begin{tabular}[b]{l c c c c c c c c}\toprule
\multirow{2}{*}{Method} & \multirow{2}{*}{\makecell{Pre-train \\ data}} & \multirow{2}{*}{\makecell{Shared \\ params}} & \multicolumn{3}{c}{WER (\%) LR} & \multicolumn{3}{c}{WER (\%) HR} \\
\cmidrule(lr){4-6} \cmidrule(lr){7-9}
 & & & V & A & AV & V & A & AV \\
\midrule\midrule
\textbf{Base(+) models} & & & & & & & & \\
AV-HuBERT~\cite{shi2022learning} & LRS3 & \xmark & 51.8 & 4.9 & 4.7 & 44.0 & 3.0 & 2.8 \\
VATLM~\cite{zhu2023vatlm} & LRS3 & \xmark & 48.0 & - & 3.6 & - & - & - \\
RAVEn~\cite{haliassos2022jointly} & LRS3 & \xmark & 47.0 & 4.7 & - & 39.1 & 2.2 & - \\
AV-data2vec~\cite{lian2023av} & LRS3 & \xmark & 45.2 & 4.4 & 4.2 & 39.0 & \underline{2.0} & \underline{1.8} \\
Lip2Vec~\cite{djilali2023lip2vec} & LRS3 & \xmark & 49.5 & - & - & 42.0 & - & - \\
BRAVEn~\cite{haliassos2024braven} & LRS3 & \xmark & \underline{43.4} & \underline{4.0} & \underline{4.0} & \underline{36.0} & \textbf{1.9} & - \\
\rowcolor{LightCyan}
USR & LRS3 & \cmark & \textbf{36.0} & \textbf{3.2} & \textbf{3.0} & \textbf{34.3} & \textbf{1.9} & \textbf{1.6} \\
\midrule
\textbf{Base(+) models} & & & & & & & & \\
AV-HuBERT~\cite{shi2022learning} & LRS3+Vox2 & \xmark & 46.1 & 4.6 & 4.0 & 34.8 & 2.0 & 1.8 \\
VATLM~\cite{zhu2023vatlm} & LRS3+Vox2 & \xmark & 42.6 & - & 3.4 & 34.2 & - & 1.7 \\
RAVEn~\cite{haliassos2022jointly} & LRS3+Vox2 & \xmark & 40.2 & 3.8 & - & 33.1 & 1.9 & - \\
AV-data2vec~\cite{lian2023av} & LRS3+Vox2 & \xmark & 37.8 & 3.7 & \underline{3.3} & 32.9 & 1.7 & \underline{1.4} \\ 
Lip2Vec~\cite{djilali2023lip2vec} & LRS3+Vox2 & \xmark & 40.6 & - & - & 34.1 & - & - \\
BRAVEn~\cite{haliassos2024braven} & LRS3+Vox2 & \xmark & \underline{35.1} & \underline{3.0} & - & \underline{28.8} & \textbf{1.4} & - \\ 
\rowcolor{LightCyan}
USR & LRS3+Vox2 & \cmark & \textbf{28.4} & \textbf{2.6} & \textbf{2.5} & \textbf{26.5} & \underline{1.6} & \textbf{1.3} \\ 
\midrule
\textbf{Large models} & & & & & & & & \\
AV-HuBERT~\cite{shi2022learning} & LRS3+Vox2 & \xmark & 32.5 & 2.9 & 3.3 & 28.6 & \underline{1.3} & 1.4 \\
VATLM~\cite{zhu2023vatlm} & LRS3+Vox2 & \xmark & 31.6 & - & \underline{2.7} & 28.4 & - & \underline{1.2} \\
RAVEn~\cite{haliassos2022jointly} & LRS3+Vox2 & \xmark & 32.5 & 2.7 & - & 28.2 & 1.4 & - \\
AV-data2vec~\cite{lian2023av} & LRS3+Vox2 & \xmark & \underline{30.8} & 2.7 & \underline{2.7} & 28.5 & \underline{1.3} & 1.3 \\
Lip2Vec~\cite{djilali2023lip2vec} & LRS3+Vox2 & \xmark & 31.2 & - & - & \underline{26.0} & - & - \\
BRAVEn~\cite{haliassos2024braven} & LRS3+Vox2 & \xmark & \underline{30.8} & \textbf{2.3} & - & 26.6 & \textbf{1.2} & - \\
u-HuBERT~\cite{hsu2022u} & LRS3+Vox2 & \cmark & - & - & - & 29.1 & 1.5 & 1.3 \\
\rowcolor{LightCyan}
USR & LRS3+Vox2 & \cmark & \textbf{26.9} & \underline{2.4} & \textbf{2.4} & \textbf{22.3} & \textbf{1.2} & \textbf{1.1}  \\ 
\bottomrule 
\end{tabular} 
\end{table}

\vspace{-0.1cm}
\subsection{Comparisons with Self-supervised Methods}
Table~\ref{table:self_supervised_lrs3} compares our approach on LRS3~\cite{afouras2018lrs3} with self-supervised methods under similar model sizes and data settings. We combine pre-training (Section~\ref{sec:self}) with standard fine-tuning (Section~\ref{sec:scratch}) when using identical pre-training and fine-tuning data, and with semi-supervised fine-tuning (Section~\ref{sec:semi}) when using extra unlabelled data. In addition to the low-resource labelled data setting outlined in Section~\ref{sec:semi_properties}, we test in a high-resource setting using the full 433-hour LRS3 dataset for fine-tuning. Our pre-training employs either LRS3 alone or combined with a 1,326-hour English-only version of VoxCeleb2~\cite{chung2018voxceleb2,shi2022learning}. We experiment with Base, Base+, and Large Transformers (see Appendix~\ref{app:model_config}).
\paragraph{Low-resource.} Using the Base model and LRS3 for pre-training, our approach significantly exceeds the previous state-of-the-art across VSR, ASR, and AVSR, when fine-tuning on 30 hours. Increasing the pre-training data and model size enhances performance, demonstrating our method's scalability. With the Large model and LRS3+Vox2 as pre-training data, we achieve 26.9\% WER for VSR and 2.4\% WER for both ASR and AVSR, matching BRAVEn on ASR and surpassing it on VSR. Unlike other methods, which use separate models for each task, \textit{USR employs a single model for all tasks}.
\paragraph{High-resource.} 
In the high-resource setting, our results are comparable to modality-specific models for ASR/AVSR and superior for VSR across all settings. Our top model obtains 22.3\% WER for VSR, 1.2\% WER for ASR, and 1.1\% WER for VSR, significantly outperforming u-HuBERT, which also uses a single model for all modalities. Furthermore, USR's low-resource VSR performance is superior to u-HuBERT's high-resource VSR result.

\begin{table}
\centering
\caption{\textbf{Comparisons with the state-of-the-art on LRS3.} \textsuperscript{*}Labels include automatic transcriptions from ASR models trained on large-scale, often non-public datasets. ``ST'' desnote offline self-training.}
\label{table:lrs3_sota}
\begin{tabular}[b]{l c c c c c c c}\toprule
\multirow{2}{*}{Method} & \multirow{2}{*}{\makecell{Labelled \\ hours}} & \multirow{2}{*}{\makecell{Unlabelled \\ hours}} & \multirow{2}{*}{\makecell{Language \\ model}} & \multirow{2}{*}{\makecell{Shared \\ params}} & \multicolumn{3}{c}{WER (\%)} \\
\cmidrule(lr){6-8}
& & & & & V & A & AV \\
\midrule\midrule
\textbf{Supervised\textsuperscript{*}} & & & & & & & \\
V2P~\cite{shillingford2018large} & 3,886 & - & \xmark & \xmark & 55.1 & - & - \\
RNN-T~\cite{makino2019recurrent} & 31,000 & - & \xmark & \cmark & 33.6 & 4.8 & 4.5 \\
VTP~\cite{prajwal2022sub} & 2,676 & - & \cmark & \xmark & 30.7 & - & - \\
Auto-AVSR~\cite{ma2023auto} & 1,902 & - & \cmark & \xmark & 23.5 & \textbf{1.0} & 1.0 \\
Auto-AVSR~\cite{ma2023auto} & 3,448 & - & \cmark & \xmark & 19.1 & \textbf{1.0} & \textbf{0.9} \\
ViT3D-CM~\cite{serdyuk2022transformer} & 90,000 & - & \xmark & \xmark & 17.0 & - & 1.6 \\
SynthVSR~\cite{liu2023synthvsr} & 6,720 & - & \cmark & \xmark & 16.9 & - & - \\
LP Conf~\cite{chang2024conformer} & 100,000 & - & \xmark & \xmark & \textbf{12.8} & - & \textbf{0.9} \\
\midrule
\textbf{Self/semi-supervised} & & & & & & & \\
AV-HuBERT w/ ST~\cite{shi2022learning} & 433 & 1,326 & \xmark & \xmark & 28.6 & - & - \\
RAVEn w/ ST~\cite{haliassos2022jointly} & 433 & 1,326 & \cmark & \xmark & 23.1 & 1.4 & - \\
\rowcolor{LightCyan}
USR & 433 & 1,326 & \cmark & \cmark & \textbf{21.5} & \textbf{1.2} & \textbf{1.1} \\
\bottomrule 
\end{tabular}
\end{table}

\begin{table}
\centering
\caption{\textbf{Comparisons with the state-of-the-art on LRS2.} \textsuperscript{*}Includes methods that use automatic transcriptions from ASR models trained on large-scale datasets. ``ST'' stands for self-training.}
\label{table:lrs2_sota}
\vspace{0.1cm}
\begin{tabular}[b]{l c c c c c c c}\toprule
\multirow{2}{*}{Method} & \multirow{2}{*}{\makecell{Labelled \\ hours}} & \multirow{2}{*}{\makecell{Unlabelled \\ hours}} & \multirow{2}{*}{\makecell{Language \\ model}} & \multirow{2}{*}{\makecell{Shared \\ params}} & \multicolumn{3}{c}{WER (\%)} \\
\cmidrule(lr){6-8}
& & & & & V & A & AV \\
\midrule\midrule
\textbf{Supervised\textsuperscript{*}} & & & & & & & \\
CM-seq2seq~\cite{ma2021end} & 380 & - & \cmark & \xmark & 37.9 & 3.9 & 3.7 \\
CM-aux~\cite{ma2022visual} & 1,459 & - & \cmark & \xmark & 25.5 & - & - \\
VTP~\cite{prajwal2022sub} & 698 & - & \cmark & \xmark & 28.9 & - & - \\
VTP~\cite{prajwal2022sub} & 2,676 & - & \cmark & \xmark & 22.6 & - & - \\
Auto-AVSR~\cite{ma2023auto} & 818 & - & \cmark & \xmark & 27.9 & 2.6 & - \\
Auto-AVSR~\cite{ma2023auto} & 3,448 & - & \cmark & \xmark & \textbf{14.6} & \textbf{1.5} & \textbf{1.5} \\
\midrule
\textbf{Self/semi-supervised} & & & & & & & \\
Uni-AVSR~\cite{pan2022leveraging} & 223 & 60,000 & \xmark & \xmark & 43.2 & 2.7 & 2.6 \\
LiRA~\cite{ma2021lira} & 223 & 433 & \cmark & \xmark & 38.8 & - & - \\
RAVEn~\cite{haliassos2022jointly} & 223 & 1,759 & \xmark & \xmark & 23.2 & 2.5 & - \\
RAVEn w/ ST~\cite{haliassos2022jointly} & 223 & 1,759 & \cmark & \xmark & 17.9 & 2.3 & - \\
\rowcolor{LightCyan}
USR & 223 & 1,759 & \xmark & \cmark & 16.0 & 2.0 & \textbf{1.9} \\
\rowcolor{LightCyan}
USR & 223 & 1,759 & \cmark & \cmark & \textbf{15.4} & \textbf{1.9} & \textbf{1.9} \\
\bottomrule 
\end{tabular}
\end{table}

\subsection{Comparisons with the State-of-the-Art} 

\paragraph{LRS3.} In Table~\ref{table:lrs3_sota}, we compare our best model against the state-of-the-art on LRS3. We present our USR results with a language model incorporated via shallow fusion~\cite{haliassos2022jointly,ma2023auto}, improving VSR performance from 22.3\% to 21.5\%. Despite using a shared model for all tasks, our performance exceeds multiple supervised methods and approaches top results~\cite{ma2023auto,serdyuk2022transformer,liu2023synthvsr}, which use significantly more labelled data. USR surpasses Auto-AVSR on VSR (21.5\% vs. 23.5\%) despite the latter using more total data and external ASR models for transcription. Finally, we outperform self-supervised methods~\cite{shi2022learning,haliassos2022jointly} using self-training that require a costly beam search strategy combining CTC, attention, and language model scores. Our simpler, greedy approach is effective, and we aim to explore additional offline pseudo-labelling for USR in future work. 

\paragraph{LRS2.} We also compare with the state-of-the-art on the LRS2 dataset~\cite{son2017lip} (see Table \ref{table:lrs2_sota}). We train our model using the same hyperparameters as for the high-resource LRS3 setting. Consistent with our LRS3 results from Table~\ref{table:lrs3_sota}, USR surpasses all other self-supervised methods across ASR, VSR, and AVSR, and outperforms strong supervised methods~\cite{ma2023auto} trained with $>$\,4$\times$ more labelled data (433 vs. 1,759 hours). Results on the WildVSR dataset are in Appendix~\ref{app:more_sota}.

\section{Conclusion} Despite their similarities, research in VSR, ASR, and AVSR has typically focused on developing separate models for each task. In this paper, we propose unified training strategies that use a single model to address all three tasks simultaneously. Our USR approach combines self-supervised learning with a greedy pseudo-labelling semi-supervised technique to achieve state-of-the-art results, surpassing related methods that use separate models for each task. Future work could explore alternative encoder architectures, strategies to improve pseudo-label quality, and methods to incorporate extra audio-only data. We hope to inspire further efforts towards consolidating ASR, VSR, and AVSR systems.

\section*{Acknowledgements}
Only Imperial College co-authors downloaded, accessed, and used the datasets. Imperial College authors conducted all of the dataset pre-processing at Imperial College.
 
{
\small

\bibliography{example_paper}
\bibliographystyle{ieee}
}

\clearpage
\appendix

\section{Limitations} \label{sec:limitations}
USR uses unlabelled samples during fine-tuning via pseudo-labelling, which is more computationally intensive than standard supervised fine-tuning due to (1) the increased data volume and (2) the high cost of pseudo-labelling. However, our semi-supervised approach \textit{without} pre-training still outperforms state-of-the-art self-supervised methods (37.8\% vs. 43.4\% WER~\cite{haliassos2024braven} in the LRS3 low-resource setting). Additionally, our approach efficiently generates pseudo-labels using a simple thresholding mechanism. Despite this, higher-quality labels are known to improve speech recognition, often enhanced by techniques like beam search, language modelling, and combining CTC and attention scores. We do not explore alternative filtering mechanisms, which we defer to future work.

\section{Societal Impact} \label{sec:societal_impact}
Speech recognition technology can greatly benefit people with disabilities who may struggle to interact with devices using traditional input methods like keyboards. Visual speech recognition can assist individuals with aphonia, who cannot produce voiced speech. It has also been shown that models trained for visual speech recognition can also aid in detecting fake videos by understanding natural mouth movements~\cite{haliassos2021lips}.

However, speech recognition technology also poses societal risks. It can be exploited for surveillance through, \textit{e.g.}, CCTV, necessitating appropriate government regulations. As in other machine learning applications, there may be biases in the datasets used to train the models. Biases related to gender, age, or ethnic background can lead to reduced performance for underrepresented groups. Addressing this requires training models on balanced data or employing bias-reduction techniques.

\section{Experiment Details} 
\subsection{Dataset Details}
\paragraph{LRS3.} The LRS3 dataset \cite{afouras2018lrs3} is the largest publicly accessible audio-visual dataset for continuous speech recognition with transcriptions. It includes approximately 430 hours of spoken sentences from TED Talks and features a vocabulary of over 50,000 words spoken by thousands of different speakers. The dataset is collected by automatically tracking faces, synchronising the video/audio streams, and splitting the videos into individual sentences. The test set comprises roughly 1 hour of utterances from speakers not included in the training set.

\paragraph{LRS2.} The LRS2 dataset \cite{son2017lip}, totalling 223 hours of footage from BBC programs, is the second-largest transcribed audio-visual dataset available for continuous speech recognition. The test set is around 0.5 hours long. Like LRS3, LRS2 features an unrestricted vocabulary and includes thousands of diverse speakers. However, LRS3 tends to contain videos of more variable quality, making it a more challenging dataset for VSR.

\paragraph{WildVSR.} WildVSR \cite{djilali2024vsr} is a recent VSR dataset, created by closely following the LRS3 dataset curation processes. The VSR dataset contains more challenging samples compared with LRS3, leading to significant drops in the VSR performance of models evaluated on WildVSR. The test set contains around 5 hours of footage.

\paragraph{VoxCeleb2.} VoxCeleb2 \cite{chung2018voxceleb2} is a large-scale audio-visual dataset containing talking faces of celebrities, with about 6,000 speakers and over 2,400 hours of footage. The dataset includes elements like laughter, cross-talk, music, and other interference, with an unconstrained vocabulary. Since VoxCeleb2 is multilingual, we use an English-only version curated by \cite{shi2022learning}, which consists of 1,323 hours of footage.

\label{sec:experiment_details}
\subsection{Data Licenses} \label{app:data_licenses}
LRS3~\cite{afouras2018lrs3}, VoxCeleb2~\cite{chung2018voxceleb2}, and WildVSR~\cite{djilali2024vsr} are licensed under CC BY-NC-ND 4.0. LRS2~\cite{son2017lip} allows for academic, non-commercial research. 

\subsection{Pre-processing}
We follow the video pre-processing protocol from related works~\cite{ma2022visual,shi2022learning,haliassos2022jointly,haliassos2024braven}. We remove motion jitter from the videos, crop a $96\times 96$ region centred around the mouth for each frame, and apply a grayscale transformation. We note that raw audio is used without pre-processing. As in~\cite{shi2022learning,haliassos2022jointly,haliassos2024braven}, we tokenise the targets using SentencePiece~\cite{kudo2018sentencepiece} subword units with a vocabulary size of 1,000.

\subsection{Model Configurations} \label{app:model_config}

\begin{table}
\centering
\caption{\textbf{Configuration of our models.} Unlike in~\cite{shi2022learning,haliassos2022jointly,haliassos2024braven}, the number of parameters includes the whole model, including the decoder and feature extractors.}
\label{table:nets_config}
\vspace{0.1cm}
\begin{tabular}[b]{l c c c}\toprule
 & Base & Base+ & Large  \\ \midrule
Parameters (M) & 86 & 171 & 503 \\ 
Encoder blocks & 12 & 12 & 24 \\
Decoder blocks & 6 & 6 & 9 \\
Attention dimension & 512 & 768 & 1024  \\ 
Attention heads & 8 & 12 & 16 \\
MLP size & 2048 & 3072 & 4096 \\ 
\bottomrule 
\end{tabular}
\end{table}

Following~\cite{haliassos2024braven}, we use three model sizes: Base, Base+, and Large. While the Transformer encoders and decoders vary in size, the feature extractors remain unchanged, consistent with \cite{haliassos2022jointly,haliassos2024braven} (which use the same feature extractors). The configuration of the models is summarised in Table~\ref{table:nets_config}. Base+ corresponds to the Base models used in similar works~\cite{shi2022learning,zhu2023vatlm,lian2023av}. We train our Base, Base+, and Large models on 32, 64, and 128 A100 40GB GPUs, respectively.

\begin{table}
\centering
\caption{\textbf{Supervised/semi-supervised training settings.}}
\label{table:sup_hparams}
\vspace{0.1cm}
\begin{tabular}[b]{l c}
\toprule
Hyperparameter & Value \\ \midrule
Training epochs & 75 \\
Warmup epochs & 20 \\
Optimiser & AdamW \\
Learning rate & 3e-3 (LRS3), 2e-3 (LRS3+Vox2) \\
Optimiser $(\beta_1,\beta_2)$ & $(0.9,0.98)$ \\
Weight decay & 0.04 \\
Learning rate schedule & Cosine decay \\
Drop rate~\cite{huang2016deep} & 0.1 (Base), 0.2 (Large)  \\
Gradient clipping threshold & 3.0 \\
Video augmentations & RandomCrop + HorizontalFlip \\ 
Frames per GPU (labelled) & 155 (low-resource), 700 (high-resource) \\
Frames per GPU (unlabelled) & 2,400 (LRS3), 1,400 (LRS3+Vox2) \\ \bottomrule
\end{tabular}
\end{table}

\subsection{Supervised/Semi-supervised Training Settings} \label{app:lab_settings}

We use consistent settings across supervised training (Section~\ref{sec:scratch}) and semi-supervised training (Section~\ref{sec:semi}). We train our models using AdamW~\cite{loshchilov2017decoupled} for 75 epochs with a 20-epoch linear warmup~\cite{goyal2017accurate} and a cosine learning rate decay~\cite{loshchilov2016sgdr}. We use gradient clipping and drop path~\cite{huang2016deep} for regularisation. In addition to the masking discussed in the main text, we also perform random spatial cropping (size $88\times 88$) and horizontal flipping (probability 0.5) on the videos in a temporally consistent manner, as in~\cite{haliassos2022jointly,haliassos2024braven}. The hyperparameter details are presented in Table~\ref{table:sup_hparams}. We fix the seed to 42. It takes approximately 12 hours to train the Base model on the labelled data (32 GPUs). It takes around one, four, and six days to train the Base (32 GPUs), Base+ (64 GPUs), and Large (128 GPUs) models, respectively. Note that Base is trained on LRS3, and Base+ and Large on LRS3+Vox2. 

\subsection{Pre-training Settings} \label{app:self_settings}

\begin{table}
\centering
\caption{\textbf{Settings for pre-training.}}
\label{table:self_hparams}
\vspace{0.1cm}
\begin{tabular}[b]{l c}
\toprule
Hyperparameter & Value \\ \midrule
Training epochs & 150 (LRS3), 75 (LRS3+VoxCeleb2) \\
Warmup epochs & 40 (LRS3), 20 (LRS3+VoxCeleb2) \\
Optimiser & AdamW \\
Learning rate & 5e-3 (LRS3), 2e-3 (LRS3+VoxCeleb2) \\
Optimiser $(\beta_1,\beta_2)$ & $(0.9,0.98)$ \\
Weight decay & 0.04 \\
Learning rate schedule & Cosine decay \\
Drop rate~\cite{huang2016deep} & 0.1 (Base), 0.2 (Large)  \\
Gradient clipping threshold & 3.0 \\
Video augmentations & RandomCrop + HorizontalFlip \\ 
Frames per GPU & 2,400 (Base), 1,800 (Base+), 900 (Large) \\ \bottomrule
\end{tabular}
\end{table}

The pre-training settings are similar. We use a longer schedule (in terms of number of epochs) for LRS3 with 150 total training epochs and 40 warmup epochs. We also use a higher learning rate of $5\times 10^{-3}$. The full settings are given in Table~\ref{table:self_hparams}. It takes approximately two days to pre-train all models. 

\subsection{Decoding}
We use the ESPNet framework~\cite{watanabe2018espnet} for decoding, as in~\cite{haliassos2022jointly,haliassos2024braven}, employing beam search with a beam size of 40. The final beam search score is
\begin{align}
 \mathcal{S}=\alpha\mathcal{S}_{\text{ctc}}+(1-\alpha)\mathcal{S}_{\text{att}}+\beta\mathcal{S}_{\text{lm}},   
\end{align}
where $\mathcal{S}_{\text{ctc}}$ and $\mathcal{S}_{\text{att}}$ are scores from the CTC and attention branches, respectively, and $\mathcal{S}_{\text{lm}}$ is the optional score from a pre-trained language model, which is incorporated through shallow fusion \cite{watanabe2017hybrid}. Following \cite{ma2023auto,haliassos2022jointly}, we set $\alpha=0.1$ for all experiments. When using a language model, we select $\beta$ from $\{0.1, 0.2, 0.3, 0.4\}$ based on the validation set. 

\section{More Ablations}

\begin{table}
\caption{\textbf{More semi-supervised ablations} under the LRS3 low-resource setting using our Base model (includes self-supervised pre-training).}
\vspace{-0.7cm}
\label{table:more_semi_ablations}
\begin{subtable}[t]{0.48\textwidth}
    \centering
    \vspace{-1.8cm}
    \caption{\textbf{Filtering thresholds} $\tau_{\text{ctc}}$ and $\tau_{\text{att}}$ for CTC and attention, respectively.}
    \label{table:more_thresholds}
    \begin{tabular}{l l c c c}\toprule
    \multirow{2}{*}{$\tau_{\text{ctc}}$} & \multirow{2}{*}{$\tau_{\text{att}}$} & \multicolumn{3}{c}{WER (\%)} \\
    \cmidrule(lr){3-5}
    & & V & A & AV \\
    \midrule
    0.60 & 0.60 & 37.2 & 3.3 & 3.1 \\
    \rowcolor{Gray}
    0.80 & 0.80 & \textbf{36.0} & 3.2 & 3.0 \\
    0.95 & 0.95 & 36.6 & 3.3 & 3.1 \\
    0.60 & 0.80 & 36.7 & \textbf{3.1} & \textbf{2.9} \\
    0.95 & 0.80 & 36.2 & 3.3 & 3.2 \\
    0.80 & 0.60 & 37.7 & 3.2 & 3.0 \\
    0.80 & 0.95 & 36.5 & 3.3 & 3.1 \\
    \bottomrule 
    \end{tabular}
\end{subtable}%
\hfill
\begin{subtable}[h]{0.48\textwidth}
\raisebox{-2cm}{
\begin{minipage}[t]{\linewidth}
\centering
\caption{\textbf{Hard versus soft sampling.}}
\label{table:soft_sampling}
\begin{tabular}[b]{l c c c}\toprule
\multirow{2}{*}{Sampling} & \multicolumn{3}{c}{WER (\%)} \\  \cmidrule(lr){2-4}
 & V & A & AV \\ \midrule
 \rowcolor{Gray}
Hard & \textbf{36.0} & \textbf{3.2} & \textbf{3.0} \\
Soft & 37.5 & 3.4 & 3.4 \\
\bottomrule 
\end{tabular}
\end{minipage}
}
\end{subtable}
\end{table}

\subsection{Semi-supervised ablations.} \label{sec:more_semi_ablations}

\paragraph{Confidence threshold.} Our default setting uses a pseudo-labelling confidence threshold $\tau$ of 0.8 both for the CTC and attention losses, for simplicity. In Table \ref{table:more_thresholds}, we investigate different threshold values, including the use of separate thresholds for the two losses. We observe that USR's performance remains consistent across a range of different thresholds, with no clear improvement when using separate thresholds.

\paragraph{Hard versus soft sampling.} Our greedy attention pseudo-labelling strategy involves choosing at each generation step the most likely pseudo-label according to the probability distribution given by the decoder. For comparison, we consider an alternative ``soft sampling'' approach as well. We use weighted sampling at each generation step, drawing a label based on the entire distribution given by the decoder. Each label has a chance of being selected proportional to its estimated probability. This approach increases the variety of pseudo-labels but may reduce their quality since low-probability pseudo-labels are more frequently used.

In Table \ref{table:soft_sampling} we compare the two approaches. We observe that hard sampling outperforms soft sampling for all three modalities. Future work can explore alternative methods to effectively increase pseudo-label variety.

\begin{table}
\caption{\textbf{More self-supervised ablations} under the LRS3 low-resource setting using our Base model.}
\label{table:more_self_sup_ablations}
\begin{subtable}[h]{0.47\textwidth}
\centering
\caption{\textbf{Mask probability.}}
\label{table:mask_prob}
\begin{tabular}[b]{l c c c}\toprule
\multirow{2}{*}{Mask probability} & \multicolumn{3}{c}{WER (\%)} \\  \cmidrule(lr){2-4}
 & V & A & AV \\ \midrule
0.2 & 37.1 & 3.4 & 3.2 \\
\rowcolor{Gray}
0.4 & \textbf{36.0} & 3.2 & 3.0 \\
0.6 & 36.7 & \textbf{3.1} & \textbf{2.9} \\
0.8 & 38.0 & 3.3 & 3.1 \\
\bottomrule 
\end{tabular}
\end{subtable}
\hfill
\begin{subtable}[h]{0.47\textwidth}
\centering
\caption{\textbf{Pre-training target types.}}
\label{table:targets_sup_ft2}
\begin{tabular}[b]{l c c c}\toprule
\multirow{2}{*}{Target} & \multicolumn{3}{c}{WER (\%)} \\  \cmidrule(lr){2-4}
 & V & A & AV \\ \midrule
AV & \textbf{36.0} & \textbf{3.2} & \textbf{3.0}\\
A+V+AV & 36.2 & \textbf{3.2} & \textbf{3.0} \\
\bottomrule 
\end{tabular}
\end{subtable}
\end{table}

\subsection{Self-supervised ablations.}

\paragraph{Mask probability.}
In Table \ref{table:mask_prob}, we compare different mask probabilities for pre-training. A low mask probability can result in a trivial learning task, whereas a high probability can make the task overly challenging. We find that a probability between 0.4 and 0.6 achieves a good balance.

\paragraph{Combining targets.}
During pre-training, targets are generated from audio-visual input and predicted by students using masked auditory, visual, and audio-visual inputs. We explore predicting the combined targets from all input types by summing the corresponding outputs from the teacher, but as shown in Table \ref{table:targets_sup_ft2}, this does not yield improvements over simply predicting the audio-visual targets.

\section{Comparisons with the State-of-the-Art on WildVSR} \label{app:more_sota}
WildVSR~\cite{djilali2024vsr} is a recent test set featuring more challenging "in-the-wild" samples than LRS3. In Table~\ref{table:wild}, we evaluate our Large model on WildVSR, trained using the high-resource setting (see Table~\ref{table:lrs3_sota}). Our unified approach achieves similar VSR results to the modality-specific RAVEn when the latter uses an additional self-training stage.

\begin{table}
\centering
\caption{\textbf{WildVSR results.} We test our model from Table \ref{table:lrs3_sota} for VSR.}
\label{table:wild}
\vspace{0.1cm}
\begin{tabular}[b]{l c c c c}\toprule
\multirow{2}{*}{Method} & \makecell{Labelled \\ hours} & \makecell{Unlabelled \\ hours} & \makecell{Shared \\ params} & WER (\%) \\
\midrule\midrule
\textbf{Supervised} & & & & \\
CM-seq2seq~\cite{ma2021end} & 1,459 & - & \xmark & 58.4 \\
VTP~\cite{prajwal2022sub} & 698 & - & \xmark & 75.6 \\
VTP~\cite{prajwal2022sub} & 2,676 & - & \xmark & 68.7 \\
Auto-AVSR~\cite{ma2023auto} & 661 & - & \xmark & 62.3 \\
Auto-AVSR~\cite{ma2023auto} & 1,759 & - & \xmark & 49.3 \\
Auto-AVSR~\cite{ma2023auto} & 3,448 & - & \xmark & 38.6 \\
\midrule
\textbf{Self/semi-supervised} & & & & \\
AV-HuBERT~\cite{shi2022learning} & 433 & 1,326 & \xmark & 51.7 \\
AV-HuBERT w/ self-training~\cite{shi2022learning} & 433 & 1,326 & \xmark & 48.7 \\
RAVEn~\cite{haliassos2022jointly} & 433 & 1,326 & \xmark & 52.2 \\
RAVEn w/ self-training~\cite{haliassos2022jointly} & 433 & 1,326 & \xmark & 46.7 \\
\rowcolor{LightCyan}
USR & 433 & 1,326 & \cmark & \textbf{46.4} \\
\bottomrule 
\end{tabular}
\end{table}

\begin{table}
\caption{\textbf{Comparisons between supervised and our semi-supervised fine-tuning.} We use the LRS3 low-resource setting and our Base model.}
\label{table:self_vs_semi_ft}
\begin{subtable}[h]{0.47\textwidth}
\centering
\caption{\textbf{Fine-tuning ``tricks'' for supervised and semi-supervised fine-tuning.}}
\label{table:ft_tricks}
\begin{tabular}[b]{l c c c}\toprule
\multirow{2}{*}{Fine-tuning} & \multicolumn{3}{c}{WER (\%)} \\  \cmidrule(lr){2-4}
 & V & A & AV \\ \midrule
Sup & 52.5 & 5.8 & 5.4 \\
Sup w/ tricks & \textbf{45.6} & \textbf{5.2} & \textbf{5.0} \\ \midrule
Semi & \textbf{36.0} & \textbf{3.2} & \textbf{3.0} \\
Semi w/ tricks & 39.3 & \textbf{3.2} & \textbf{3.0} \\
\bottomrule 
\end{tabular}
\end{subtable}
\hfill
\begin{subtable}[h]{0.47\textwidth}
\centering
\caption{\textbf{Pre-training target types for supervised fine-tuning.}}
\label{table:targets_sup_ft}
\begin{tabular}[b]{l c c c}\toprule
\multirow{2}{*}{Target} & \multicolumn{3}{c}{WER (\%)} \\  \cmidrule(lr){2-4}
 & V & A & AV \\ \midrule
V & 63.2 & 9.0 & 8.9 \\
A & \textbf{43.9} & \textbf{4.8} & \textbf{4.6 }\\
AV & 45.6 & 5.2 & 5.0 \\
\bottomrule 
\end{tabular}
\end{subtable}
\end{table}

\section{Supervised vs. Semi-supervised Fine-tuning} \label{app:sup_vs_semi}

In Table~\ref{table:self_vs_semi_ft}, we closely evaluate the differences between supervised and semi-supervised fine-tuning.

Supervised fine-tuning with few labelled samples is prone to overfitting, necessitating various training ``tricks'' to improve performance. For example, \cite{haliassos2022jointly,haliassos2024braven} use a smaller decoder for the low-resource setting, different learning rates for the encoder and decoder, and layer-wise learning rate decay~\cite{clark2020electra}. We use our Base model and the low-resource setting to evaluate supervised and semi-supervised (our default) fine-tuning, with and without these strategies. As shown in Table~\ref{table:ft_tricks}, while these ``tricks'' significantly benefit supervised training (consistent with \cite{haliassos2022jointly}), they actually \textit{hurt} semi-supervised fine-tuning. This suggests that semi-supervised training is less prone to overfitting, making these regularisation methods unnecessary. In general, we noticed that using semi-supervised fine-tuning results in less sensitivity to pre-training hyperparameters (\textit{e.g.}, compare Tables \ref{table:targets_sup_ft} and \ref{table:target_type}).

In Table~\ref{table:target_type}, we observed that our semi-supervised fine-tuning benefits most from audiovisual targets. Here, we fine-tune the same pre-trained model using only labeled data to assess the influence of target type on supervised fine-tuning. Table~\ref{table:targets_sup_ft} shows that audio-only targets perform best for supervised fine-tuning, consistent with findings from other works~\cite{lian2023av,haliassos2022jointly}. As discussed in the main text, semi-supervised fine-tuning allows the model to leverage the rich and diverse information in audiovisual targets, which supervised fine-tuning struggles to achieve.

\begin{table}
\centering
\caption{\textbf{Experiments with auditory noise.} We compare USR with the modality-specific BRAVEn method on LRS3 with different signal-to-noise-ratio (SNR) levels. We use Base models trained under the low-resource setting.}
\label{table:av_results}
\vspace{0.1cm}
\begin{tabular}[b]{l l l l l}\toprule
 & & \multicolumn{3}{c}{SNR (dB)} \\ \cmidrule(lr){3-5}
 & \multicolumn{1}{c}{Clean} & \multicolumn{1}{c}{5} & \multicolumn{1}{c}{0} & \multicolumn{1}{c}{-5} \\ \midrule
BRAVEn (A) & \textbf{4.0} & 15.6 & 24.6 & \,\,\,99.0  \\ 
BRAVEn (AV) & \textbf{4.0} \textcolor{Gray}{$\scriptstyle = $} & \textbf{12.4} \textcolor{ForestGreen}{$\scriptstyle \downarrow 3.2$} & \textbf{15.0} \textcolor{ForestGreen}{$\scriptstyle \downarrow 9.6$} & \,\,\,\textbf{48.5} \textcolor{ForestGreen}{$\scriptstyle \downarrow 50.5$} \\ \midrule
USR (A) & 3.2 & 14.3 & 26.9 & 100.4 \\
USR (AV) & \textbf{3.0} \textcolor{ForestGreen}{$\scriptstyle \downarrow \mathbf{0.2}$} & \,\,\,\textbf{6.1} \textcolor{ForestGreen}{$\scriptstyle \downarrow \mathbf{8.2}$} & \textbf{10.1} \textcolor{ForestGreen}{$\scriptstyle \downarrow \mathbf{16.8}$} & \,\,\,\textbf{35.7} \textcolor{ForestGreen}{$\scriptstyle \downarrow \mathbf{64.7}$}  \\
\bottomrule 
\end{tabular}
\end{table}

\begin{table}
\centering
\caption{\textbf{Error bars.} We report the mean and standard deviation over five runs with random seeds. We use our Base model with LRS3 as the pre-training dataset.}
\label{table:error_bars}
\vspace{0.1cm}
\begin{tabular}[b]{l c c c}\toprule
\multirow{2}{*}{Setting} & \multicolumn{3}{c}{WER (\%)} \\  \cmidrule(lr){2-4}
 & V & A & AV \\ \midrule
Low-resource & $36.2 \pm 0.40$ & $3.25 \pm 0.10$ & $3.02 \pm 0.04$ \\
High-resource & $34.2 \pm 0.56$ & $1.77 \pm 0.18$ & $1.68 \pm 0.11$ \\
\bottomrule 
\end{tabular}
\end{table}

\section{Experiments with Auditory Noise}
We have demonstrated that AVSR slightly outperforms ASR on the clean LRS3 test set. However, it is in the presence of auditory noise that AVSR truly excels, as visual cues help clarify ambiguous utterances. Table~\ref{table:av_results} presents ASR and AVSR results under varying levels of audio babble noise from the NOISEX dataset~\cite{varga1993assessment}. We employ our Base model under the low-resource setting with LRS3 as the pre-training dataset. Notably, the noise is added to the LRS3 test set, and the model is not trained on noisy data. We observe that as noise levels increase (and the signal-to-noise ratio decreases), the performance gap between AVSR and ASR widens. Interestingly, this gap is more pronounced for USR compared to the modality-specific BRAVEn. 

\section{Error Bars} \label{sec:error_bars}
Due to high computational demands and in line with previous studies~\cite{shi2022learning,haliassos2022jointly,haliassos2024braven,lian2023av,hsu2022u}, we do not include error bars for our main results. To assess the variability of our method across multiple training runs, Table~\ref{table:error_bars} presents the mean and standard deviation over five runs with different random seeds for our low- and high-resource settings, using our Base model with LRS3 as the pre-training dataset. We observe that the results are consistently stable around the mean.

\section{Qualitative Differences between Self-supervised Pretext Tasks}
Our pre-training method shares similarities with recent audio-visual self-supervised tasks, RAVEn~\cite{haliassos2022jointly}, BRAVEn~\cite{haliassos2024braven}, and AV-data2vec~\cite{lian2023av}. These methods employ an EMA-based teacher to generate targets from unmasked data, which the student predicts using masked inputs. Here, we compare and contrast our USR pretext task with these methods. 

\subsection{Comparisons with RAVEn/BRAVEn} RAVEn and BRAVEn pre-train separate Transformer encoders for visual and auditory inputs, which are then fine-tuned for ASR and VSR. AVSR can be performed through shallow fusion of visual and auditory features. In contrast, USR pre-trains a single student Transformer encoder for auditory, visual, and audiovisual inputs, significantly reducing training and inference costs.

We adopt the approach of using a shallow Transformer encoder as a predictor, which has been shown to improve representation learning~\cite{haliassos2022jointly}. However, while RAVEn and BRAVEn use separate predictors for visual and auditory features (with BRAVEn also using differently-sized predictors), we use a single predictor for all modalities, simplifying the architectural design.

\subsection{Comparisons with AV-data2vec}
AV-data2vec also unifies pre-training by using a single Transformer encoder for all modalities. However, while AV-data2vec employs random modality sampling, we compute all per-modality losses at each iteration, amortising the cost of target generation (see Section~\ref{sec:scratch_properties}). AV-data2vec's use of a scheduler for modality probabilities increases the complexity of the pre-training process. Furthermore, AV-data2vec uses audio-only targets, whereas we use audiovisual targets, which are shown to perform best for our semi-supervised fine-tuning (see Section~\ref{sec:semi}).

\section{Comparison with AV-CPL} \label{app:avcpl_comparison}

\begin{table}
\centering
\caption{\textbf{Comparison with AV-CPL.} LRS3 results for the low-resource (LR) and high-resource (HR) labelled data settings. We show results for the Large model using LRS3+Vox2 as the pre-training dataset.}
\label{table:usr_vs_avcpl}
\vspace{0.1cm}
\begin{tabular}[b]{l c c c c c c}\toprule
\multirow{2}{*}{Method} & \multicolumn{3}{c}{WER (\%) LR} & \multicolumn{3}{c}{WER (\%) HR} \\
\cmidrule(lr){2-4} \cmidrule(lr){5-7}
 & V & A & AV & V & A & AV \\
\midrule
AV-CPL \cite{rouditchenko2023av} & 56.7 & 10.0 & 10.4 & 47.4 & 2.3 & 2.2 \\
USR & \textbf{26.9} & \textbf{2.4} & \textbf{2.4} & \textbf{22.3} & \textbf{1.2} & \textbf{1.1}  \\ 
\bottomrule 
\end{tabular} 
\end{table}

As mentioned in Section~\ref{sec:related_work}, the recent AV-CPL method~\cite{rouditchenko2023av} uses pseudo-labelling to train a single model for ASR, VSR, and AVSR, similar to our semi-supervised approach described in Section~\ref{sec:semi}. Table~\ref{table:usr_vs_avcpl} compares USR with AV-CPL on the low- and high-resource labelled data settings using the Large model and LRS3+Vox2 as the pre-training dataset. We observe dramatic WER differences between the two methods, which we attribute to USR's use of CTC-attention training, self-supervised pre-training, and pseudo-label filtering, among other design choices studied in Section~\ref{sec:main_properties}.

\section{Summary of the Impact of Semi- and Self-supervised Training}
Sections \ref{sec:semi_properties}, \ref{sec:self_properties}, and Appendix \ref{app:sup_vs_semi} demonstrate the impact of self- and semi-supervised learning on speech recognition performance. Table \ref{table:summary_main_contributions} summarizes the contributions of each component. Self-supervised pre-training on the full LRS3 dataset, followed by supervised fine-tuning on 30 hours of LRS3 (see Appendix \ref{app:sup_vs_semi}), outperforms supervised training on the same 30 hours alone, as expected. Additionally, semi-supervised training (without pre-training) significantly surpasses the self-supervised baseline. Combining self-supervised pre-training with semi-supervised fine-tuning yields the best results.

\begin{table}
\centering
\caption{\textbf{Summary of the impact of semi- and self-supervised training} under the LRS3 low-resource setting using our Base model. We compare four approaches: supervised training on 30 hours of labelled data, self-supervised pre-training with supervised fine-tuning, semi-supervised training, and self-supervised pre-training with semi-supervised fine-tuning.}
\label{table:summary_main_contributions}
\vspace{0.1cm}
\begin{tabular}[b]{l c c c c c}\toprule
\multirow{2}{*}{Setting} & \multirow{2}{*}{{\makecell{Self-supervised \\ pre-training}}} & \multirow{2}{*}{Fine-tuning} & \multicolumn{3}{c}{WER (\%)} \\  \cmidrule(lr){4-6}
 & & & V & A & AV \\ \midrule
Only labelled data & \xmark & Supervised & 61.8 & 8.9 & 8.4 \\
Self-supervised & \cmark & Supervised & 43.9 & 4.8 & 4.6 \\ 
Semi-supervised & \xmark & Semi-supervised & 37.8 & 4.0 & 3.9 \\
Self- + semi-supervised & \cmark & Semi-supervised & \textbf{36.0} & \textbf{3.2} & \textbf{3.0} \\
\bottomrule 
\end{tabular}
\end{table}

\begin{table}
\caption{\textbf{Failure cases on the LRS3 test set}. We use the Large model trained in the high-resource setting with LRS3+VoxCeleb2.}
\vspace{0.1cm}
\label{table:transcriptions}
\centering
\resizebox{\linewidth}{!}{
\begin{tabular}[b]{l c}\toprule
Source & Transcription \\
\cmidrule(lr){1-2}
Groundtruth & And all of this matters greatly because public safety to me is the most important function \\
VSR &  \textcolor{ForestGreen}{And all of} \textcolor{red}{these matters are crazy} \textcolor{ForestGreen}{because public safety to me is the most important function} \\
ASR & \textcolor{ForestGreen}{And all of this matters greatly because public safety to me is the most important function} \\
AVSR & \textcolor{ForestGreen}{And all of this matters greatly because public safety to me is the most important function} \\
\midrule
Groundtruth & I'm here to tell you the story of crazy love, a psychological trap disguised as love \\
VSR &  \textcolor{ForestGreen}{I'm here to tell you the story of crazy love, a psychological trap} \textcolor{red}{denies the} \textcolor{ForestGreen}{love} \\
ASR & \textcolor{ForestGreen}{I'm here to tell you the story of crazy love, a psychological trap disguised as love} \\
AVSR & \textcolor{ForestGreen}{I'm here to tell you the story of crazy love, a psychological trap disguised as love} \\
\midrule
Groundtruth & It took six days to deploy a global malware campaign \\
VSR & \textcolor{ForestGreen}{It took six days to deploy our global} \textcolor{red}{market} \textcolor{ForestGreen}{campaign} \\
ASR & \textcolor{ForestGreen}{It took six days to deploy our global} \textcolor{red}{Mali Wear} \textcolor{ForestGreen}{campaign} \\
AVSR & \textcolor{ForestGreen}{It took six days to deploy a global malware campaign} \\
\midrule
Groundtruth & It worked for the Oakland A's and it worked in the state of New Jersey \\
VSR &  \textcolor{ForestGreen}{It worked for the} \textcolor{red}{Oaklands} \textcolor{ForestGreen}{and it worked in the state of New Jersey} \\
ASR & \textcolor{ForestGreen}{It worked for the Oakland} \textcolor{red}{Asia} \textcolor{ForestGreen}{and it worked in the state of New Jersey} \\
AVSR & \textcolor{ForestGreen}{It worked for the Oakland A's and it worked in the state of New Jersey} \\
\bottomrule 
\end{tabular}
}
\end{table}

\section{Failure Cases} Table \ref{table:transcriptions} presents some failure cases from the LRS3 test set. We evaluated our Large model trained in a high-resource setting with LRS3 and VoxCeleb2. While VSR tends to produce more errors than ASR and AVSR, these errors are often related to phonetically similar sounds, such as ``this'' vs. ``these'' or ``disguised'' vs. ``denies.'' Additionally, using both auditory and visual modalities (AVSR) can improve the model's ability to distinguish challenging samples, such as ``Mali Wear'' vs. ``malware.''

\newpage
\section*{NeurIPS Paper Checklist}

\begin{enumerate}

\item {\bf Claims}
    \item[] Question: Do the main claims made in the abstract and introduction accurately reflect the paper's contributions and scope?
    \item[] Answer: \answerYes{} % Replace by \answerYes{}, \answerNo{}, or \answerNA{}.
    \item[] Justification: See Sections~\ref{sec:intro},~\ref{sec:method}, and~\ref{sec:sota_comparisons}.

\item {\bf Limitations}
    \item[] Question: Does the paper discuss the limitations of the work performed by the authors?
    \item[] Answer: \answerYes{} % Replace by \answerYes{}, \answerNo{}, or \answerNA{}.
    \item[] Justification: See Appendix~\ref{sec:limitations}.

\item {\bf Theory Assumptions and Proofs}
    \item[] Question: For each theoretical result, does the paper provide the full set of assumptions and a complete (and correct) proof?
    \item[] Answer: \answerNA{} % Replace by \answerYes{}, \answerNo{}, or \answerNA{}.
    \item[] Justification: The paper does not include theoretical results.

    \item {\bf Experimental Result Reproducibility}
    \item[] Question: Does the paper fully disclose all the information needed to reproduce the main experimental results of the paper to the extent that it affects the main claims and/or conclusions of the paper (regardless of whether the code and data are provided or not)?
    \item[] Answer: \answerYes{} % Replace by \answerYes{}, \answerNo{}, or \answerNA{}.
    \item[] Justification: See Section \ref{sec:main_properties} and Appendix \ref{sec:experiment_details}.

\item {\bf Open access to data and code}
    \item[] Question: Does the paper provide open access to the data and code, with sufficient instructions to faithfully reproduce the main experimental results, as described in supplemental material?
    \item[] Answer: \answerYes{} % Replace by \answerYes{}, \answerNo{}, or \answerNA{}.
    \item[] Justification: We provide code as part of the supplementary material.

\item {\bf Experimental Setting/Details}
    \item[] Question: Does the paper specify all the training and test details (e.g., data splits, hyperparameters, how they were chosen, type of optimizer, etc.) necessary to understand the results?
    \item[] Answer: \answerYes{} % Replace by \answerYes{}, \answerNo{}, or \answerNA{}.
    \item[] Justification: See Section \ref{sec:main_properties} and Appendix \ref{sec:experiment_details}.

\item {\bf Experiment Statistical Significance}
    \item[] Question: Does the paper report error bars suitably and correctly defined or other appropriate information about the statistical significance of the experiments?
    \item[] Answer: \answerYes{} % Replace by \answerYes{}, \answerNo{}, or \answerNA{}.
    \item[] Justification: See Appendix~\ref{sec:error_bars}.

\item {\bf Experiments Compute Resources}
    \item[] Question: For each experiment, does the paper provide sufficient information on the computer resources (type of compute workers, memory, time of execution) needed to reproduce the experiments?
    \item[] Answer: \answerYes{} % Replace by \answerYes{}, \answerNo{}, or \answerNA{}.
    \item[] Justification: See Appendices~\ref{app:model_config},~\ref{app:lab_settings}, and~\ref{app:self_settings}.
    
\item {\bf Code Of Ethics}
    \item[] Question: Does the research conducted in the paper conform, in every respect, with the NeurIPS Code of Ethics \url{https://neurips.cc/public/EthicsGuidelines}?
    \item[] Answer: \answerYes{} % Replace by \answerYes{}, \answerNo{}, or \answerNA{}.
    \item[] Justification: The research does not violate the NeurIPS Code of Ethics.

\item {\bf Broader Impacts}
    \item[] Question: Does the paper discuss both potential positive societal impacts and negative societal impacts of the work performed?
    \item[] Answer: \answerYes{} % Replace by \answerYes{}, \answerNo{}, or \answerNA{}.
    \item[] Justification: See Appendix~\ref{sec:societal_impact}.
    
\item {\bf Safeguards}
    \item[] Question: Does the paper describe safeguards that have been put in place for responsible release of data or models that have a high risk for misuse (e.g., pretrained language models, image generators, or scraped datasets)?
    \item[] Answer: \answerNA{} % Replace by \answerYes{}, \answerNo{}, or \answerNA{}.
    \item[] Justification: The paper poses no such risks.

\item {\bf Licenses for existing assets}
    \item[] Question: Are the creators or original owners of assets (e.g., code, data, models), used in the paper, properly credited and are the license and terms of use explicitly mentioned and properly respected?
    \item[] Answer: \answerYes{} % Replace by \answerYes{}, \answerNo{}, or \answerNA{}.
    \item[] Justification: See Section~\ref{app:data_licenses}.

\item {\bf New Assets}
    \item[] Question: Are new assets introduced in the paper well documented and is the documentation provided alongside the assets?
    \item[] Answer: \answerNA{} % Replace by \answerYes{}, \answerNo{}, or \answerNA{}.
    \item[] Justification: We do not release new assets.

\item {\bf Crowdsourcing and Research with Human Subjects}
    \item[] Question: For crowdsourcing experiments and research with human subjects, does the paper include the full text of instructions given to participants and screenshots, if applicable, as well as details about compensation (if any)? 
    \item[] Answer: \answerNA{} % Replace by \answerYes{}, \answerNo{}, or \answerNA{}.
    \item[] Justification: The paper does not involve crowdsourcing or research with human subjects.

\item {\bf Institutional Review Board (IRB) Approvals or Equivalent for Research with Human Subjects}
    \item[] Question: Does the paper describe potential risks incurred by study participants, whether such risks were disclosed to the subjects, and whether Institutional Review Board (IRB) approvals (or an equivalent approval/review based on the requirements of your country or institution) were obtained?
    \item[] Answer: \answerNA{} % Replace by \answerYes{}, \answerNo{}, or \answerNA{}.
    \item[] Justification: The paper does not involve crowdsourcing or research with human subjects.

\end{enumerate}

\end{document}